\documentclass[sigconf]{acmart}

\AtBeginDocument{%
  \providecommand\BibTeX{{%
    \normalfont B\kern-0.5em{\scshape i\kern-0.25em b}\kern-0.8em\TeX}}}

\setcopyright{acmcopyright}
\copyrightyear{2023}
\acmYear{2023}
\acmDOI{XXXXXXX.XXXXXXX}

\acmConference[Preprint]{Preprint. Under Review.}{Under Review}{}
\acmPrice{15.00}
\acmISBN{978-1-4503-XXXX-X/18/06}

\begin{document}

\title{GASCOM: Graph-based Attentive Semantic Context Modeling for Online Conversation Understanding}


\author{Vibhor Agarwal}
\affiliation{%
  \institution{University of Surrey}
  \city{Guildford}
  \state{Surrey}
  \country{UK}}
\email{v.agarwal@surrey.ac.uk}

\author{Yu Chen}
\affiliation{%
  \institution{Meta AI}
  \city{Menlo Park}
  \state{California}
  \country{USA}}
\email{hugochen@meta.com}

\author{Nishanth Sastry}
\affiliation{%
  \institution{University of Surrey}
  \city{Guildford}
  \state{Surrey}
  \country{UK}}
\email{n.sastry@surrey.ac.uk}






\renewcommand{\shortauthors}{Agarwal, et al.}
\newcommand{\hugo}[1]{{\color{red}{\%Hugo: #1}}}
\begin{abstract}
  Online conversation understanding is an important yet challenging NLP problem which has many useful applications (e.g., hate speech detection). However, online conversations typically unfold over a series of posts and replies to those posts, forming a tree structure within which individual posts may refer to semantic context from higher up the tree. Such semantic cross-referencing makes it difficult to understand a single post by itself; yet considering the entire conversation tree is not only difficult to scale but can also be misleading as a single conversation may have several distinct threads or points, not all of which are relevant to the post being considered. 
  In this paper, we propose a \textbf{G}raph-based \textbf{A}ttentive \textbf{S}emantic \textbf{CO}ntext \textbf{M}odeling (GASCOM) framework for online conversation understanding. Specifically, we design two novel algorithms that utilise both the graph structure of the online conversation as well as the semantic information from individual posts for retrieving relevant context nodes from the whole conversation. We further design a \textit{token-level} multi-head graph attention mechanism to pay different attentions to different tokens from different selected context utterances for fine-grained conversation context modeling. Using this semantic conversational context, we re-examine two well-studied problems: polarity prediction and hate speech detection. Our proposed framework significantly outperforms state-of-the-art methods on both tasks, improving macro-F1 scores by $4.5\%$ for polarity prediction and by $5\%$ for hate speech detection. The GASCOM context weights also enhance interpretability.
\end{abstract}

\ccsdesc[500]{Computing methodologies~Natural language processing}
\ccsdesc[500]{Computing methodologies~Discourse, dialogue and pragmatics}
\ccsdesc[300]{Computing methodologies~Machine learning algorithms}
\ccsdesc[500]{Information systems~World Wide Web}

\keywords{graph-based models, online conversation understanding, hate speech, polarity prediction}


\maketitle

\section{Introduction}\label{sec:introduction}

The Internet has enabled people to participate in sharing their views freely and indulge in online conversations. \textcolor{black}{Online conversation understanding is an important yet challenging NLP problem which has many useful applications such as hate speech~\cite{fortuna2018survey,agarwal2021hate,ghosh2023cosyn} and polarity prediction~\cite{agarwal2022graphnli,cocarascu2020dataset,agarwal2022graph}}. Similar to conventional threaded conversations~\cite{reddy2018coqa,choi2018quac}, an utterance in an online conversation might refer back to its conversation history via coreference or ellipsis; hence effectively utilizing the conversation context is a key ingredient to understanding online conversations. Unlike conventional conversations which are usually represented as a sequence of alternating utterances from multiple speakers, online conversations usually explicitly form a tree structure where each node represents a post or comment and the edges are directed from a given node to the other (parent) node it is replying to~\cite{agarwal2022graphnli}. Therefore, it is also important to model this graph structure~\cite{chen2020graphflow,ghosal2019dialoguegcn}.

One effective approach to utilizing such conversation context is to augment the semantic meanings of the target utterance with relevant utterances retrieved from other posts in the same conversation. This process involves two steps: \textit{i)} utilizing the graph structure of the conversation tree to select  relevant utterance nodes for sampling, and \textit{ii)} selecting the parts of the sampled utterances which are most relevant to the target utterance. 

To sample relevant nodes (Step \textit{i}), unlike previous works on conversation modeling which only utilize either the semantic meanings of the conversation context~\citep{bordes2016learning,kim2017speaker} or its graph structure~\citep{agarwal2022graphnli}, in this work we propose \emph{novel semantic-aware random walk algorithms} for retrieving relevant context nodes from the conversation where the transition probability between context nodes is determined by the strength of their semantic relationship. Both algorithms use random walks to explore surrounding graph structure for context, but select the nodes differently. The first method samples each possible next step in the random walk with a probability proportional to the cosine similarity between S-BERT embeddings of the text of that post with the post being considered. The second method uses attention instead of similarity, and selects the next node in each step of the random walk with a probability proportional to its attention weight. Once the nodes (posts) representing the relevant semantic context are selected,  
 instead of treating the importance of each selected utterance equally, 
we propose to learn different attention weights to different tokens from different selected context utterances via a further \emph{token-level multi-head graph attention mechanism} when aggregating their semantic meanings (Step \textit{ii)}.

We highlight our contributions as follows:
\begin{itemize}
    \item We propose a general deep learning framework --- GASCOM --- for online conversation understanding by effectively utilizing conversational context to augment the semantic meanings of the target utterance.
    \item We design two novel \textit{semantic-aware graph-based conversation context selection algorithms} for retrieving relevant context nodes from an online conversation which consider both the graph structure and semantic meanings of the conversation context. 
    \item We design a \textit{token-level multi-head graph attention mechanism} to pay different attentions to different tokens from different selected context utterances for fine-grained conversation context modeling.
    \item We show that our proposed framework significantly outperforms state-of-the-art methods on two very important online conversation understanding tasks, including polarity prediction and hate speech detection by $4.5\%$ and $5\%$ in macro-F1, respectively. Experimental results also show that our proposed framework has good interpretability.
\end{itemize}


\section{Related Work} \label{sec:related-work}

\subsection{Online Conversation Understanding}

Online conversation understanding is an important yet challenging NLP problem which has many useful applications (e.g., hate speech detection). \textcolor{black}{Like conventional threaded conversations, context of an utterance might refer to previous utterances in online conversations. But instead of alternating sequence of utterances, online conversations usually form a tree structure where a comment can get multiple replies.
Previous works have either utilized the semantic meanings of the conversation context~\cite{ashraf2021abusive,bordes2016learning,kim2017speaker} or its tree structure~\cite{agarwal2022graphnli} to sample the relevant utterances.
\citet{ashraf2021abusive} used replies to short Youtube comments as additional conversational context for abusive language detection.}
\textcolor{black}{\citet{agarwal2022graphnli} introduced GraphNLI that utilizes the tree structure to capture the conversational context through a biased root-seeking random walk. It selects the conversational context probabilistically with a probability $p$ without looking at the semantic meanings. Furthermore, it uses weighted average aggregation to discount and aggregate the conversational context with a discount factor $\gamma$. These hyperparameters depend upon the dataset and therefore, needs to be fine-tuned. Moreover, the discount factor gives exponentially decreasing weights as the graph walk moves away from the target comment node, which may not be always true and depends upon the context of the neighbouring nodes such as ancestors and sibling nodes.} 

\textcolor{black}{Unlike previous works, our proposed model utilizes both the semantic meaning and tree-structure to capture the conversational context. We propose novel semantic-aware random walk algorithms for retrieving relevant context nodes from the conversation which are driven by the strength of their semantic relationship. Furthermore, we propose a multi-head graph attention mechanism to learn different attention weights to different tokens from different selected context utterances when aggregating their semantic meanings.}

\subsubsection{\textbf{Polarity Prediction}} \label{sec:polarity-prediction}

Polarity prediction aims to identify the argumentative relations of \textit{attack} and \textit{support} between natural language arguments in online debates and conversations wherein one comment replies to the other comment. The polarity prediction task is one of the many tasks in the field of \textit{argument mining} (e.g. \citet{lawrence2020argument,lippi2016argumentation,cabrio2018five}).
The polarity prediction task is important because it helps to deduce the stance of a comment with respect to the other. Once we have classified all the replies in a debate, we can apply ideas from argumentation theory to reason about which arguments should be justified. \textit{Argumentation theory} is a branch of AI that is concerned with the transparent and rational resolution of disagreements (e.g. \citet{rahwan2009argumentation}).


The polarity prediction task has been discussed in the literature. For example, \citet{cabrio2018five} has reviewed the task in the context of persuasive essays or political debates. An early example of this work is \citet{cabrio2013natural}, which applied textual entailment (e.g. \citet{bos2006logical,dagan2010recognizing,maccartney2008modeling}) to predict the polarity of replies on the now-defunct Debatepedia dataset. In \citet{cocarascu2017identifying}, long-short-term memory networks were used to classify polarities. A more recent overview of the polarity prediction task \cite{cocarascu2020dataset} has provided context-independent neural network baselines.


\subsubsection{\textbf{Hate Speech Detection}} \label{sec:hate-speech-detection}

Internet debates, especially those about controversial topics, can easily spread hate and misinformation~\cite{cinelli2021online,guest2021expert,jahan2021systematic}. Hate speech is notoriously difficult to define. A sample of important attempted definitions (e.g. \citet{jahan2021systematic}) agree that hate speech is a public language that attacks individuals and groups of people because of protected characteristics, for example, their race, skin colour, religion, ancestry, nationality, gender, disability, sexuality and so on. Hate speech, if left unchallenged, can promote and incite harmful societal consequences against individuals and groups such as (but not limited to) physical attacks, psychological intimidation, property damage, violence and segregation. Therefore, it is important to be able to detect hate speech in online forums, accurately and at scale, such that appropriate action can be taken by the moderators.

\begin{figure}[h]
  \centering
  \includegraphics[width=\linewidth]{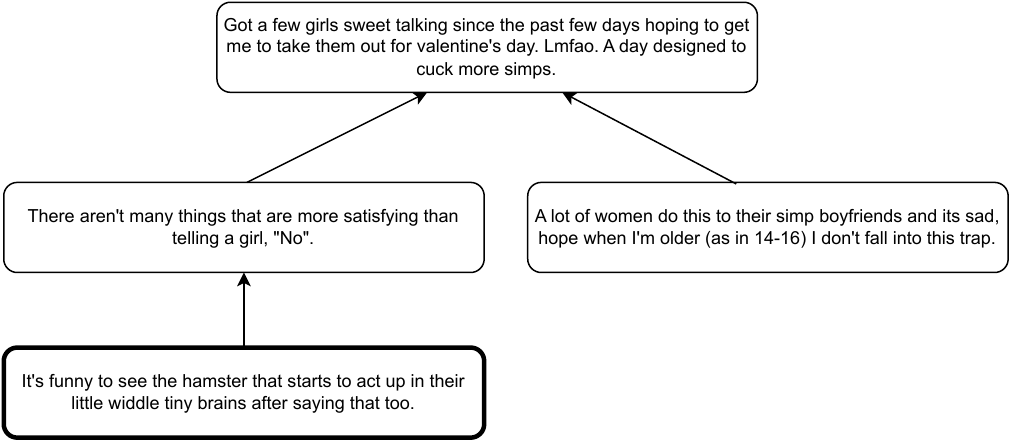}
  \caption{An example conversation from \textit{Guest} dataset. (\textbf{Warning: Contains misogynistic speech)}}
  \label{fig:guest-reddit-hate}
\end{figure}

Figure~\ref{fig:guest-reddit-hate} shows an example conversation. The rectangular text boxes represent posts, and the arrows denote which posts reply to which other posts. Suppose we wish to identify whether the text with the thicker border in the bottom left contains hate speech. At first glance, this text appears to mention that hamsters have tiny brains and as such does not appear to be hate speech. However, upon looking at the neighbouring comments, we can notice that the conversation is actually about women. Therefore, this comment is misogynistic as hamsters actually refer to women. Examples like this demonstrate that although hate exists and should be dealt with accordingly, the accurate detection of hate speech is very important.

\subsection{Graph Machine Learning in NLP}
Many NLP problems can be boiled down to graph-based problems in the end. Classical graph-based algorithms have been successfully applied to numerous NLP applications~\citep{mihalcea2011graph}. For example, random walk algorithms have been applied to query expansion~\citep{collins2005query} and keyword extraction~\citep{mihalcea2004textrank} \textcolor{black}{where the node pair similarity is measured by the probability scores in the stationary distribution of the random walk on a graph.}
Graph clustering algorithms have been applied to solve text clustering~\citep{erkan2006language} where a graph of document nodes is constructed to capture the relationships among documents. Label propagation algorithms have been applied to word-sense disambiguation~\citep{niu2005word} and sentiment analysis~\citep{goldberg2006seeing} by propagating labels from limited labeled nodes to a large amount of similar unlabeled nodes with the assumption of like attracts like.

In the past few years, Graph Neural Networks (GNNs)~\citep{kipf2016semi,velivckovic2017graph,hamilton2017inductive,chen2020iterative} have drawn great attention as a special class of neural networks which can model arbitrary graph-structured data. GNNs have been widely applied to various NLP tasks such as machine translation~\cite{DBLP:conf/emnlp/BastingsTAMS17}, code summarization~\citep{liu2021retrieval}, natural question generation~\citep{chen2020reinforcement,chen2020toward} and machine reading comprehension~\citep{chen2020graphflow}.
We refer the interested reader to the comprehensive survey by~\cite{wu2021graph} for more details.

\section{GASCOM Architecture} \label{sec:methodology}

In this section, we propose our general GASCOM architecture for online conversation understanding which leverages both the semantic meaning of the conversation context and the tree structure of the conversations. In Section~\ref{discussion-trees}, we discuss how we represent online conversations as discussion trees. In Section~\ref{context-select}, we propose semantic-aware graph-based algorithms for conversation context selection followed by our model architecture and token-level multi-head graph attention for online conversation understanding in Section~\ref{graphnli-attn}.

\begin{figure*}[h]
  \centering
  \includegraphics[width=\linewidth]{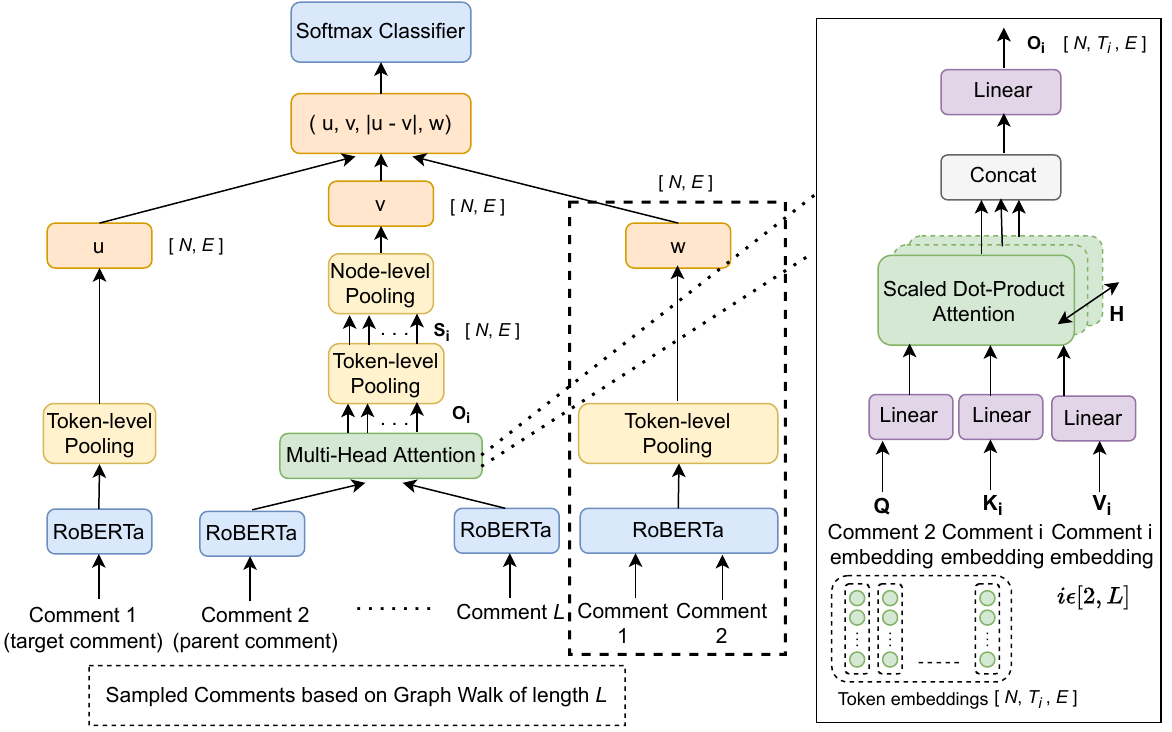}
  \caption{GASCOM architecture. Token-level Multi-head Graph Attention (Section~\ref{graphnli-attn}) is computed between comment 2 (parent comment) and all the other neighbouring comments sampled through graph walks (Section~\ref{context-select}). The token-level attention outputs $\vec{O}_i$ are mean-pooled to get their corresponding sentence embeddings $\vec{S}_i$ which are then aggregated through node-level mean pooling to get the resultant embedding $\vec{v}$. $N$, $T_i$, $E$ represent batch size, target sequence length and dimension size, respectively. Dashed box is optional and is only employed in polarity prediction task, but not in hate speech detection.}
  \label{fig:graphnli-attn-arch}
\end{figure*}

\subsection{Online Conversations as Discussion Trees} \label{discussion-trees}

For every online discussion D, we construct a discussion tree, where a node represents a post or comment and the edges are directed from a given node to the other (parent) node it is replying to. The discussion forms a tree structure because it starts with a root node, which represents the opening comment or post of the discussion. Every non-root node replies to exactly one other node (out-degree = 1), while all nodes can have zero or more replies to it (in-degree $\in$ N). Each such node has an associated label depending upon the prediction task. For polarity prediction, the non-root nodes are labelled with support or attack, depending upon whether the post is respectively for or against its parent post. For misogynistic hate speech, each node is labelled as either hate or non-hate.

\subsection{Graph-based Semantic Context} \label{context-select}

Graph walk is a principled way of capturing the conversational context for a given comment in a discussion tree for online conversation understanding~\cite{agarwal2022graphnli}. \textcolor{black}{Previous works either did not sample the relevant conversational context at all~\cite{ashraf2021abusive,bordes2016learning} or sampled it probabilistically~\cite{agarwal2022graph} without looking at the semantics.}
In this section, we propose our semantic-aware random walk strategies that select conversation context through semantic relevance of neighbouring nodes with respect to the target comment node.

\subsubsection{\textbf{Similarity-based Random Walk}} \label{sim-based-walk}

Similarity-based Random Walk is a walk that starts from a target comment node and uses similarity scores to sample the neighbouring nodes in a discussion tree probabilistically. It is a semantic-aware walk and can select relevant neighbouring nodes to sample the conversation context.
The similarity score is computed using a pre-trained Sentence-BERT model with cosine similarity~\cite{reimers-2019-sentence-bert} between a pair of nodes, \textcolor{black}{as widely used in the literature~\cite{bithel2021unsupervised,santander2022semantic,alfarizy2022verification}. It works well in our use case. However, our approach is agnostic to the choice of text similarity metric}. The resultant similarity score is normalised by dividing every score with the sum of all the scores to get their corresponding probability values (between 0 and 1) that sum to 1. Starting from a target comment node, these similarity scores and their corresponding probabilities are calculated for every one-hop neighbouring node directly connected with the target node. Then the random walk selects one of the nodes according to their assigned probabilities and therefore, the walk is non-deterministic. Likewise, the walk continues to move towards neighbours of the neighbours and so on till it achieves a walk-length of \textit{L} nodes or reaches one of the root or leaf nodes. \textit{L} is the maximum number of distinct nodes sampled by a graph walk including the starting node and therefore, the walk-length is ($L - 1$). It is important to limit the walk length by defining \textit{L} because online conversations can grow at a large scale, and capturing far away nodes through random walks can lead to the over-smoothing problem~\cite{chen2020measuring}. Since the walk is random, it is possible that the same node is visited more than once. In this case, we just consider the distinct nodes, ignoring the duplicates to continue the walk until \textit{L} distinct nodes are traversed or no new untraversed node is left.

Besides the aforementioned similarity-based random walk, we also explore its deterministic version called \textbf{similarity-based graph walk} that selects a neighbouring node with the highest similarity score and then moves towards the neighbours of the selected neighbouring node and so on till the walk-length $L - 1$.

\subsubsection{\textbf{Attention-modulated Random Walk}} \label{attn-mod-walk}
Attention-modulated Random Walk is a walk that starts from a target comment node and uses attention weights from the trained GASCOM model via self-distillation \cite{hinton2015distilling,zhang2019your,liu2021few} to sample the neighbouring nodes in a discussion tree. Starting from a target comment node, attention weights are obtained for each of the one-hop neighbouring nodes from the Multi-head Graph Attention layer (see Section~\ref{graphnli-attn}) of the trained GASCOM model with respect to the target node and then their corresponding probability values (between 0 and 1) are calculated by normalising the weights with the sum of all the attention weights. The resultant probability values sum to 1 so that the non-deterministic random walk selects one of the nodes according to their assigned probabilities. Likewise, the walk continues to move towards neighbours of neighbours and so on till it achieves a walk-length $L - 1$.

Besides the aforementioned attention-modulated random walk, we also explore its deterministic version called \textbf{attention-modulated graph walk} that selects a neighbouring node with the highest attention weight and then moves towards the neighbours of the selected neighbouring node and so on till the walk-length $L - 1$.

\subsection{Token-level Multi-Head Graph Attention} \label{graphnli-attn}



GASCOM is an attention-based deep learning architecture which uses token-level multi-head graph attention to find the relevant conversational context useful for understanding online conversations. The GASCOM architecture is illustrated in Figure~\ref{fig:graphnli-attn-arch}. At first, \textit{L} comments (nodes) are sampled through one of the proposed semantic-aware graph walk algorithms. Then these comments are input into the RoBERTa~\cite{DBLP:journals/corr/abs-1907-11692} model to get their corresponding token-level embeddings. Let $\vec{E}_i$ be the token-level embeddings for comment $i$ obtained from RoBERTa, where $i \in [1, L]$. The mean pooling operation is performed on token-level embeddings $\vec{E}_1$ of the target comment (comment 1) to derive its corresponding fixed-sized sentence embedding $\vec{u}$ as shown in equation~\ref{eq1}.

\begin{equation} \label{eq1}
    \vec{u} = MeanPool(\vec{E}_1)
\end{equation}


\textcolor{black}{Next we have conversational context embeddings $E_i$, where $i \in [2, L]$ from which we need to find the relevant context for online conversation understanding.} We propose the token-level multi-head graph attention mechanism to pay different attentions to different tokens from different selected context utterances for fine-grained conversation context modeling. The embedding vector $\vec{v}$ denotes the aggregated conversational context in Figure~\ref{fig:graphnli-attn-arch}. To find the relevant context nodes for the given target comment node, token-level embeddings of the surrounding nodes, including the parent node (comment 2), are input into the multi-head attention layer. Multi-head attention~\cite{vaswani2017attention} is computed at the token-level between comment 2 (parent comment) and all the other neighbouring comments that are input into the model. \textcolor{black}{We apply multi-head attention with the parent node and not the target node because our preliminary experiments show that the parent node is the most important context node. Therefore, we start random walks from the parent node (see Section~\ref{sec:parent-nodes-improvement}) and augment the semantic embedding of the parent node by selecting relevant context nodes and keeping the target node embedding as it is.}
When applying the multi-head attention mechanism, we set the query matrix $\vec{Q}$ to the comment 2 embedding $\vec{E}_2$, both key matrix $\vec{K}_i$ and value matrix $\vec{V}_i$ to the comment i embedding $\vec{E}_i$ as shown in equation~\ref{eq2}.

\begin{equation} \label{eq2}
    \vec{Q} = \vec{E}_2;
    \vec{K}_i = \vec{E}_i;
    \vec{V}_i = \vec{E}_i
\end{equation}

These $\vec{Q}$, $\vec{K}_i$, and $\vec{V}_i$ are input into the multi-head attention ($h$ attention heads) as shown in equations~\ref{eq9}, \ref{eq3} and \ref{eq4}, which learns attention weights for each of the $L - 1$ comments according to their relevance and returns their corresponding attention outputs. Note that $\vec{W}_j^Q$, $\vec{W}_j^{K}$ and $\vec{W}_j^{V}$ are linear projection weight matrices, and $d_k$ is a scaling factor which is set to the dimension size of $\vec{W}_j^{K}$.

\begin{equation} \label{eq9}
    Attention(\vec{Q}, \vec{K}, \vec{V}) = softmax(\frac{\vec{Q}\vec{K}^T}{\sqrt{d_k}}) \vec{V}
\end{equation}

\begin{equation} \label{eq3}
    \texttt{head}^i_j = Attention(\vec{Q} \vec{W}_j^Q, \vec{K}_i \vec{W}_j^{K}, \vec{V}_i \vec{W}_j^{V})
\end{equation}


\begin{equation} \label{eq4}
    MultiHead(\vec{Q}, \vec{K}_i, \vec{V}_i) = Concat(\{\texttt{head}^i_j\}_{j=1}^h)
\end{equation}

The final representation $\vec{O}_i$ for a comment $i$ is the dot product of multi-head attention output $MultiHead(\vec{Q}, \vec{K}_i, \vec{V}_i)$ with linearly projected matrix $\vec{W}^{O}$ as in equation~\ref{eq8}. Mean pooling operation is applied to all the token-level attention outputs $\vec{O}_i$ to get their corresponding fixed-length sentence embeddings $\vec{S}_i$ as in equation~\ref{eq5}. Then, the mean pooling operation is performed again on these $L -1$ sentence embeddings to get a resultant embedding $\vec{v}$ which represents the conversational context as in equation~\ref{eq6}.

\begin{equation} \label{eq8}
    \vec{O}_i = MultiHead(\vec{Q}, \vec{K}_i, \vec{V}_i) \vec{W}^{O}
\end{equation}

\begin{equation} \label{eq5}
    \vec{S}_i = MeanPool(\vec{O}_i)
\end{equation}

\begin{equation} \label{eq6}
    \vec{v} = MeanPool([\vec{S}_2, \vec{S}_3, ..., \vec{S}_L])
\end{equation}

For polarity prediction, cross-attention between the target (comment 1) and parent (comment 2) is also employed (as shown in the optional dashed box in Figure~\ref{fig:graphnli-attn-arch}) to further improve the performance. The concatenation of comment 1 and comment 2, separated by the $[SEP]$ token, is input into the RoBERTa model to get the resultant token-level embedding $\vec{E}_{1,2}$ 
which after mean pooling becomes sentence embedding $\vec{w}$ as shown in equation~\ref{eq7}. 

\begin{equation} \label{eq7}
    \vec{w} = MeanPool(\vec{E}_{1,2})
\end{equation}

Once we have embeddings $\vec{u}$, $\vec{v}$ and $\vec{w}$ (optional), we calculate an element-wise difference vector $|\vec{u}-\vec{v}|$. We then concatenate all four vectors $\vec{u}$, $\vec{v}$, $|\vec{u}-\vec{v}|$ and $\vec{w}$ together to get the final embedding vector, which is then fed into a softmax layer for the downstream prediction task.


\section{Experiments and Results} \label{sec:experiments}

\subsection{Data, Baselines and Evaluation Metrics}\label{sec:datasets}

\begin{table*}[h]
\caption{Class frequencies and percentages for each dataset. ``Positive'' refers to \textit{supportive} replies in Kialo, and the \textit{presence} of misogynistic hate speech in the Guest dataset. ``Negative'' refers to \textit{attacking} replies in Kialo, and the \textit{absence} of misogynistic hate speech.}
\label{tab:dataset-stats}
\begin{tabular}{c|ccccc}
\hline
\textbf{Task}  & \textbf{Dataset} & \textbf{Positive class} & \textbf{Negative class} & \textbf{Positive class} $\%$ & \textbf{Negative class $\%$} \\
\hline
Polarity prediction   & Kialo   & $139,722$      & $184,651$      & $43.1$              & $56.9$    \\
Hate speech detection & Guest  & $699$          & $5,868$        & $10.6$              & $89.4$    \\
\hline
\end{tabular}
\end{table*}

\noindent\textbf{Kialo dataset.}
We use the publicly available Kialo dataset for polarity prediction task. The Kialo dataset contains data from $1,560$ discussions hosted on Kialo, a debating platform as used by \citet{boschi2021has,young2021ranking,young2022modelling,agarwal2022graphnli}. Each reply in the debate is clearly labelled as attacking (negative) or supporting (positive). Table~\ref{tab:dataset-stats} shows the class frequencies.

\noindent\textbf{Guest dataset.}
We use the publicly available Guest dataset previously compiled by \citet{guest2021expert} for hate speech detection task. It is an expert-annotated, hate speech dataset, sourced from Reddit. This dataset looks at the specific type of hate against women - misogyny. Therefore, the positive class is ``misogynistic'' and negative class is ``non-misogynistic''. Table~\ref{tab:dataset-stats} shows the class frequencies.

For both the datasets, discussions have a tree structure with a root node that represents the start of the conversation.
For polarity (hate speech) prediction, every reply is either a support (hate) or attack (non-hate). We randomly sample $80\%$ of the instances into the training set with the remainder $20\%$ serving as the test set for both the datasets.

\begin{table*}
\centering
\caption{Performance (in \%) on \textit{Kialo} dataset for polarity prediction.}
\label{tab:perf-kialo}
\begin{tabular}{l|ccccc}
\hline
\textbf{Model} & \textbf{Accuracy} & \textbf{macro-F1} & \textbf{Precision} & \textbf{Recall} & \textbf{PR AUC} \\
\hline
Bag-of-Words + Logistic Regression & 67.00 & 62.00 & 62.00 & 62.00 & 65.91 \\
Sentence-BERT with classification layer & 79.86 & 75.81 & 77.86 & 73.86 & 79.32    \\
BERT: Root-seeking Graph Walk + MLP & 70.27 & 52.32 & 44.87 & 64.12 & 55.71   \\
\textcolor{black}{Graph Convolutional Networks} & 57.94 & 57.82 & 67.74 & 50.44 & 60.83 \\
GraphNLI: Graph Walk + Weighted Avg. & 81.97 & 76.89 & 76.83 & 76.96 & 80.34 \\
GraphNLI: Random Walk + Weighted Avg. & 81.95 & 78.96 & 78.94 & 78.99 & 81.90 \\
\hline
GASCOM: Random Walk & 82.01 & 81.62 & 81.68 & 81.57 & 84.69 \\
GASCOM: Similarity-based Graph Walk & 83.49 & 83.15 & 83.17 & 83.13 & 86.23 \\
GASCOM: Similarity-based Random Walk & \textbf{83.73} & 83.42 & 83.45 & 83.39 & 86.73 \\
GASCOM: Attention-modulated Graph Walk & 83.59 & 83.25 & 83.27 & 83.23 & 86.59 \\
GASCOM: Attention-modulated Random Walk & 83.71 & \textbf{83.46} & \textbf{83.52} & \textbf{83.41} & \textbf{86.89} \\
\hline
\end{tabular}
\end{table*}

\noindent\textbf{Baselines.}
We compare GASCOM with the relevant baselines including Bag-of-Words + Logistic Regression, Sentence-BERT with classification layer~\cite{reimers-2019-sentence-bert}, BERT~\cite{devlin2019bert} with root-seeking graph walk + MLP~\cite{agarwal2022graphnli}, Graph Convolutional Networks (GCNs)~\cite{kipf2016semi} and GraphNLI~\cite{agarwal2022graphnli}. \textcolor{black}{We use two-layered GCN for node classification with S-BERT~\cite{reimers-2019-sentence-bert} to obtain sentence embeddings for each node. GraphNLI captures the conversational (global) context of the conversations through probabilistic root-seeking random walk without looking at the semantics.}

\noindent\textbf{Evaluation Metrics.}
We use Accuracy, Macro-F1, Precision and Recall as evaluation metrics. Given the class imbalance nature of Guest dataset, we also use PR AUC score which is area under the precision-recall (PR) curve.


\subsection{Model Settings}
We use a batch size of $8$, Adam optimizer with learning rate $2\times 10^{-5}$, $h=5$ attention heads, walk-length $L=6$ and a linear learning rate warm-up over $10\%$ of the training data. We make GASCOM model end-to-end trainable by minimizing the cross-entropy loss computed based on the model predictions and ground-truth labels. We implement the model using Transformers~\cite{wolf-etal-2020-transformers} and PyTorch~\cite{Paszke_PyTorch_An_Imperative_2019} libraries and train it for $4$ epochs.

\subsection{Experimental Results}

Tables~\ref{tab:perf-kialo} and \ref{tab:perf-guest} show the performance of GASCOM model and various baselines on the \textit{Kialo} dataset for polarity prediction and \textit{Guest} dataset for hate speech detection, respectively. GASCOM model performs significantly better than all the baseline models in macro-F1 and PR AUC scores. 
\textcolor{black}{Graph Convolutional Networks (GCNs) have performed significantly poorer than GASCOM and other baselines due to the incorporation of too much potentially irrelevant and noisy conversational contexts. Specifically, GCNs incorporate all the nodes (broader context) within a given neighborhood (1-hop and 2-hop in our case) of the node being classified, which is not as effective as semantic-aware conversational context selection for capturing deeper context by GASCOM. Furthermore, too much nearby context may result in very noisy node embeddings, which weakens the model’s predictive ability.}
GASCOM with a biased root-seeking random walk performs about $3$ and $3.5$ percentage points better in macro-F1 than GraphNLI with the same random walk for polarity prediction and hate speech detection, respectively. This shows the better ability of multi-head graph attention in GASCOM to find the relevant conversational context as compared to the weighted average strategy in GraphNLI. Similarity-based and Attention-modulated random walks perform better than their corresponding deterministic graph walks. Non-deterministic random walk helps in selecting more diverse set of neighbouring comments instead of always the most relevant ones according to similarity scores or attention weights. The similarity-based random walk with GASCOM gives an overall macro-F1 score of $83.42\%$ for polarity prediction, which is about $4.5$ percentage points higher than GraphNLI with random walk and weighted aggregation and a macro-F1 score of $78.69\%$ for hate speech detection, which is about $4$ percentage points higher. Attention-modulated random walk with GASCOM slightly outperforms the similarity-based random walk with an overall macro-F1 of $83.46\%$ and $80.03\%$ for polarity prediction and hate speech detection respectively. \textcolor{black}{Our discussion trees are relatively small with a small walk length which limits the memory and time footprint for conversation context selection using random walks. A Kialo discussion tree has a mean of 204 nodes (arguments).}

\begin{table*}
\centering
\caption{Performance (in \%) on \textit{Guest} dataset for hate speech detection.}
\label{tab:perf-guest}
\begin{tabular}{l|ccccc}
\hline
\textbf{Model} & \textbf{Accuracy} & \textbf{macro-F1} & \textbf{Precision} & \textbf{Recall} & \textbf{PR AUC} \\
\hline
Bag-of-Words + Logistic Regression & 92.08 & 61.45 & 56.98 & 71.49 & 44.84 \\
Sentence-BERT with classification layer & 92.28 & 68.79 & 74.96 & 65.47 & 54.34 \\
BERT: Root-seeking Graph Walk + MLP & 92.14 & 66.56 & 61.89 & 71.65 & 51.27 \\
\textcolor{black}{Graph Convolutional Networks} & 92.17 & 67.11 & 62.44 & 72.45 & 52.35  \\
GraphNLI: Graph Walk + Weighted Avg. & 93.06 & 73.56 & 80.63 & 69.48 & 57.84 \\
GraphNLI: Random Walk + Weighted Avg. & 93.18 & 74.79 & 80.89 & 70.90 & 59.98 \\
\hline
GASCOM: Random Walk & 93.57 & 78.35 & 78.60 & 78.10 & 66.68 \\
GASCOM: Similarity-based Graph Walk & 92.95 & 75.59 & 79.50 & 72.79 & 60.63 \\
GASCOM: Similarity-based Random Walk & 93.73 & 78.69 & \textbf{83.69} & 75.23 & 66.89 \\
GASCOM: Attention-modulated Graph Walk & 92.63 & 75.50 & 76.95 & 74.25 & 60.57    \\
GASCOM: Attention-modulated Random Walk & \textbf{94.14} & \textbf{80.03} & 81.45 & \textbf{78.75} & \textbf{67.19}    \\
\hline
\end{tabular}
\end{table*}

\begin{table}
\centering
\caption{Impact of walk length \textit{L} on GASCOM model (performance in \%).}
\label{tab:ablation-walk-len}
\begin{tabular}{c|cccc}
\hline
\textbf{\textit{L}} & \textbf{Accuracy} & \textbf{macro-F1} & \textbf{P} & \textbf{R}\\
\hline
\multicolumn{5}{l}{Polarity Prediction}     \\
\hline
4 & 81.71 & 81.33 & 81.38 & 81.29    \\
6 & 81.91 & 81.51 & \textbf{81.58} & 81.45    \\
10 & \textbf{81.93} & \textbf{81.53} & \textbf{81.58} & \textbf{81.49}    \\
\hline
\multicolumn{5}{l}{Hate Speech Detection}    \\
\hline
4 & 91.38 & 76.12 & 77.96 & 74.60   \\
6 & \textbf{93.57} & \textbf{78.35} & 78.60 & \textbf{78.10}    \\
10 & 91.77 & 76.81 & \textbf{79.37} & 74.82    \\
\hline
\end{tabular}
\end{table}

\subsection{Hyperparameter Analysis}


Table~\ref{tab:ablation-walk-len} shows the effect of walk length \textit{L} on the performance of GASCOM model. For polarity prediction, walk length $L=6$ performs better than $L=4$ whereas walk length of $6$ and $10$ perform similarly with $L=10$ performing slightly better. Since there is no significant performance gain as opposed to the increased complexity of processing $10$ sentences, we choose $L=6$ to be the optimal walk length. For hate speech detection, clearly the walk length $L=6$ performs the best. On the other hand, \citet{agarwal2022graphnli} found $L=4$ to be the optimal walk length for GraphNLI. Therefore, semantic-aware graph walks and attention mechanism in GASCOM allows to input larger conversational context.

\subsection{Ablation Studies}
In this section, we discuss various ablation studies to understand different components of GASCOM model and their contributions.

\subsubsection{\textbf{Comparison of Token-level Multi-head Graph Attention with Others}}

Table~\ref{tab:perf-attention-strategies} shows the performance comparison of token-level multi-head graph attention on GASCOM model with others for the polarity prediction task. First is the average aggregation that weighs all the sampled neighbouring comments equally and computes average of these embeddings to get the resultant embedding $v$. Sentence-level multi-head attention is for the sentence-level embeddings of the neighbouring comments with mean pooling to get the resultant embedding $v$.
Both of these aggregation strategies perform similarly in terms of their macro-F1 score with a slightly better performance of sentence-level attention. Our proposed token-level multi-head graph attention performs significantly better with an overall gain of about $4$ percentage points in macro-F1 as compared to the two baselines strategies.

\begin{table}
\centering
\caption{Performance (in \%) comparison of token-level multi-head graph attention with other strategies on GASCOM for polarity prediction.}
\label{tab:perf-attention-strategies}
\begin{tabular}{l|cc}
\hline
\textbf{Strategy} & \textbf{Accuracy} & \textbf{macro-F1} \\ 
\hline
Average & 79.97 & 76.39 \\  
Sentence-level & 77.38 & 76.90 \\  
Token-level & \textbf{80.62} & \textbf{80.15} \\  
\hline
\end{tabular}
\end{table}

\subsubsection{\textbf{Comparison of Semantic-aware Conversation Context Selection with Naive Strategies}}

\begin{table}
\centering
\caption{Performance (in \%) comparison of proposed semantic-aware Graph Walk strategies with naive strategies for GASCOM model.}
\label{tab:perf-naive-strategies}
\begin{tabular}{l|cc}
\hline
\textbf{Strategy} & \textbf{Accuracy} & \textbf{macro-F1} \\ 
\hline
\multicolumn{3}{l}{Polarity Prediction} \\
\hline
Parent-Child nodes & 83.48 & 83.12 \\ 
Random 6 2-hop nbds & 83.55 & 83.24 \\ 
Biased Root-seeking \\ Random Walk & 82.01 & 81.62 \\ 
Similarity-based top 6 \\ 2-hop nbds & 83.53 & 83.22 \\ 
\textbf{Similarity-based} \\ \textbf{Random Walk} & \textbf{83.73} & 83.42 \\ 
\textbf{Attn-modulated} \\ \textbf{Random Walk} & 83.71 & \textbf{83.46} \\ 
\hline
\multicolumn{3}{l}{Hate Speech Detection}   \\
\hline
Parent-Child nodes & 93.18 & 75.93 \\ 
Random 6 2-hop nbds & 92.95 & 75.36 \\ 
Biased Root-seeking \\ Random Walk & 93.57 & 78.35 \\ 
Similarity-based top 6 \\ 2-hop nbds & 92.98 & 76.85 \\ 
\textbf{Similarity-based} \\ \textbf{Random Walk} & 93.73 & 78.69 \\ 
\textbf{Attn-modulated} \\ \textbf{Random Walk} & \textbf{94.14} & \textbf{80.03} \\ 
\hline
\end{tabular}
\end{table}

Our semantic-aware conversation context selection strategies -- similarity-based and attention-modulated random walks are compared with naive strategies. The first naive strategy is parent-child nodes which uses just the parent comment as additional context for the target comment. Second is random 6 two-hop neighbours strategy which selects 6 nodes ($L=6$) randomly from two-hop neighbors for a target node. Third strategy is biased root-seeking random walk~\cite{agarwal2022graphnli} which is a non-deterministic walk that selects neighbouring nodes probabilistically by setting the probability $p$ hyperparameter. Fourth is similarity-based top 6 strategy, a naive semantic-aware strategy which selects top 6 most similar nodes based on the cosine similarity from two-hop neighbours of the target comment.
Table~\ref{tab:perf-naive-strategies} shows the performance comparisons. For both tasks, our semantic-aware strategies perform better than naive as well as root-seeking random walk strategies with attention-modulated random walk performing the best.








\subsection{Model Analysis}

In this section, we show that cross-attention improves polarity prediction in Section~\ref{sec:cross-attn}. We also show that semantic-aware graph walks starting at parent nodes improve the model performance in Section~\ref{sec:parent-nodes-improvement}.

\begin{table}
\centering
\caption{Impact of embedding vector \textit{w} obtained using cross-attention between the target and parent comments in GASCOM with root-seeking random walk for polarity prediction (performance in \%).}
\label{tab:cross-attn-w}
\begin{tabular}{l|cc}
\hline
\textbf{Strategy} & \textbf{Accuracy} & \textbf{macro-F1} \\ 
\hline
Without \textit{w} & 80.62 & 80.15 \\ 
With \textit{w} & \textbf{82.01} & \textbf{81.62} \\ 
\hline
\end{tabular}
\end{table}

\begin{table}
\centering
\caption{Impact of Semantic-aware Graph Walks starting from the parent node on the performance of the GASCOM model (performance in \%).}
\label{tab:graph-walks-parent}
\begin{tabular}{l|cc}
\hline
\textbf{Model} & \textbf{macro-F1} \\ 
\hline
\multicolumn{2}{l}{Polarity Prediction} \\
\hline
Similarity-based Graph Walk & 78.90 \\ 
Similarity-based Graph Walk  \\
starting at parent node & 83.15 \\ 
Similarity-based Random Walk & 78.96 \\ 
Similarity-based Random Walk \\
starting at parent node & \textbf{83.42} \\ 
\hline
\multicolumn{2}{l}{Hate Speech Detection}   \\
\hline
Similarity-based Graph Walk & 72.01 \\ 
Similarity-based Graph Walk \\
starting at parent node & 75.59 \\ 
Similarity-based Random Walk & 76.20 \\ 
Similarity-based Random Walk \\
starting at parent node & \textbf{78.69} \\ 
\hline
\end{tabular}
\end{table}

\subsubsection{\textbf{Cross-Attention Improves Polarity Prediction}}\label{sec:cross-attn}

Table~\ref{tab:cross-attn-w} clearly shows the improvement of $1.5$ percentage points in macro-F1 in predicting the polarities on \textit{Kialo} dataset with cross-attention embedding \textit{w} in GASCOM model.

\subsubsection{\textbf{Semantic-aware Graph Walks Starting at Parent Nodes Improve Performance}}\label{sec:parent-nodes-improvement}

Table~\ref{tab:graph-walks-parent} shows the performance comparison of semantic-aware graph walks starting from the parent node versus the target node. For both tasks, conversation context selection strategies perform significantly better when they start from the parent node. In online conversations, generally the target comment (reply) uses the context of immediate parent comment and therefore replies to it directly. Furthermore, in polarity prediction, polarity of the target node is with respect to the parent node. Hence, deterministic inclusion of parent nodes in random walks is better than non-deterministic inclusion in case of semantic-aware random walk starting from the target node itself.

\subsection{Case Study and Interpretability Analysis}

\begin{figure}[h]
  \centering
  \includegraphics[width=\linewidth]{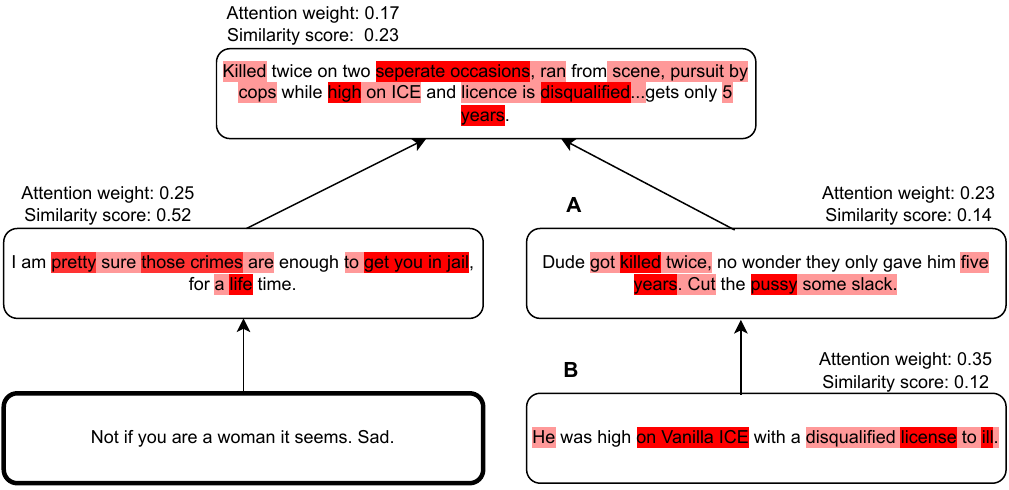}
  \caption{An example conversation from \textit{Guest} dataset. Bolded (bottom left) node is a hate speech (misogyny). Every neighbouring node has attention weight and similarity score obtained with its sentence-level embeddings. Highlighted part of the context indicates token-level attention weights. Darker the shade, higher are the attention weights.}
  \label{fig:conv-attn-highlight}
\end{figure}

For interpretability analysis of the semantic-aware graph walks and token-level multi-head graph attention, we consider an example conversation from \textit{Guest} dataset, as shown in Figure~\ref{fig:conv-attn-highlight}. Bolded (bottom left) node is actually a misogynistic hate speech. Baseline models such as BERT can not accurately predict it as hate without the conversational context, whereas our GASCOM model can accurately predict it as hate. Every neighbouring node in the conversation has attention weight (Section~\ref{attn-mod-walk}) and similarity score (Section~\ref{sim-based-walk}) obtained by comparing its sentence-level embeddings with the parent embeddings of the target node. As seen in Figure~\ref{fig:conv-attn-highlight}, token-level graph attention spreads attention weights across the neighbouring nodes and gives higher weights to the relevant nodes. For example, graph attention gives higher weights to the sibling nodes A and B in the parallel thread because they talk about women with some explicit misogynistic words. But similarity-based approach just gives higher weights to the immediate parent and ancestor nodes instead of actually considering the semantic relevance. 

Besides node-level analysis, we look at the token-level attention to understand which tokens from the selected nodes receive higher weights. Highlighted part of the context indicates token-level attention weights. Darker the red shade, higher are the attention weights. We notice that important keywords or words related to women are actually given higher attention weights, especially for the sibling nodes in the right-side thread. \textcolor{black}{This example illustrates the utility of proposed semantic-aware graph-based conversation context selection algorithms and token-level multi-head graph attention in providing appropriate weights to different nodes and different parts of texts within nodes.}




\section{Conclusions and Future Work} \label{sec:conclusions}


We proposed a general deep learning framework for online conversation understanding.
Our proposed framework employs novel semantic-aware graph-based conversation context selection algorithms and token-level multi-head graph attention mechanism to effectively utilize conversational context to augment the semantic meanings of the target utterance. We demonstrated the superiority of the proposed approach on two important online conversation understanding tasks including polarity prediction and hate speech detection. Future directions include jointly training the graph-based conversation context selection module and the remaining modules for improved performance.

\section{Limitations}

In forums such as BBC’s \textit{Have Your Say?}\footnote{\url{https://www.bbc.co.uk/blogs/haveyoursay/archives.html}}, there is no explicit threaded reply structure, requiring us to infer from the text of a reply which other post it is replying to, to construct discussion trees. In this less restrictive user interface, a single post may refer to or reply to multiple other posts, creating more than one edge and a conversation structure that is no longer a tree but a more general graph. We believe that the GASCOM model would work in this more general context as well, with the random walk sampling all the available conversational context. However, this has not been tested empirically. 

GASCOM makes use of conversation context from surrounding posts. While GASCOM can label each post as a conversation evolves and new posts are added, it becomes more effective only after a reasonable number of replies have been added. At the beginning of a conversation, when not a lot of conversation context is available, GASCOM will likely only perform similar to baseline models such as BERT, which also operate without the additional context.

Currently, GASCOM is trained on English conversations. However, with widespread multilingual online conversations and conversations in low resource languages, there is a need to build models for online conversation understanding in multilingual and low resource settings. It is easy to adapt GASCOM for low resource and multilingual settings by using language models trained on specific langauges as encoders instead of RoBERTa trained on English corpus.

\section{Ethical Considerations}

\begin{itemize}
    \item \textbf{Confidentiality:} Access to data is critical to the effectiveness of our work. We only use publicly available de-identified datasets.
    \item \textbf{Fairness and Biases:} AI models trained on huge amounts of data can have the potential for various kinds of biases associated due to the dataset and the proposed model.
    \item \textbf{Potential for Harm:} AI models are not $100\%$ accurate. In case of hate speech detection, false positives (falsely labelled hate comment) and false negatives (undetected hate) can be an issue in social media platforms. Any prediction by GASCOM of hate speech will need to be manually verified. Our multi-head graph attention mechanism can aid this process through a more nuanced interpretation of the results (as demonstrated in  Figure~\ref{fig:conv-attn-highlight}).
\end{itemize}


\bibliographystyle{ACM-Reference-Format}
\bibliography{sample-base}

\end{document}